\documentclass[runningheads]{llncs}
\usepackage{graphicx}

\usepackage[utf8]{inputenc} 
\usepackage[T1]{fontenc}    
\usepackage{hyperref}       
\usepackage{url}            
\usepackage{booktabs}       
\usepackage{amsfonts}       
\usepackage{nicefrac}       
\usepackage{microtype}      
\usepackage{multirow}
\usepackage{bbm}
\usepackage{graphicx}
\graphicspath{{./figures/}}
\usepackage{amsmath}
\usepackage{color}

\begin{document}
\title{Single-Path NAS: Designing Hardware-Efficient ConvNets in less than 4 Hours}
\titlerunning{Single-Path NAS}
%
\author{Dimitrios Stamoulis\inst{1} \and
Ruizhou Ding\inst{1} \and
Di Wang\inst{2} \and
Dimitrios Lymberopoulos\inst{2} \and
Bodhi Priyantha\inst{2} \and
Jie Liu\inst{3} \and
Diana Marculescu\inst{1}}
\authorrunning{D. Stamoulis et al.}
%
\institute{Department of ECE, Carnegie Mellon University, Pittsburgh, PA, USA \and
Microsoft, Redmond, WA, USA \and
Harbin Institute of Technology, Harbin, China\\
\email{dstamoul@andrew.cmu.edu}}

\maketitle              
\begin{abstract}
Can we automatically design a Convolutional Network (ConvNet) with the
highest image classification accuracy under the latency constraint 
of a mobile device? Neural architecture search (NAS) has revolutionized
the design of hardware-efficient ConvNets by automating this process.
However, the NAS problem remains challenging due to the combinatorially large 
design space, causing a significant searching time (at least 200 GPU-hours). 
To alleviate this complexity, we propose \textit{Single-Path NAS}, a novel 
differentiable NAS method for designing hardware-efficient ConvNets 
in \textbf{less than 4 hours}.  Our contributions are as follows: 
1.~\textbf{Single-path search space}: Compared to previous differentiable 
NAS methods, \textit{Single-Path NAS} uses one single-path over-parameterized 
ConvNet to encode all architectural decisions with shared convolutional
kernel parameters, hence drastically decreasing the number of
trainable parameters and the search cost down to few epochs. 
2.~\textbf{Hardware-efficient ImageNet classification}: 
\textit{Single-Path NAS} achieves $74.96\%$ top-1 accuracy on 
ImageNet with 79ms latency on a Pixel 1 phone, which is 
state-of-the-art accuracy compared to NAS methods 
with similar inference latency constraints ($\leq 80ms$). 3.~\textbf{NAS efficiency}: 
\textit{Single-Path NAS} search cost is only 
\textbf{8 epochs} (30 TPU-hours), which is up to \textbf{5,000$\times$ faster}
compared to prior work. 4.~\textbf{Reproducibility}:
Unlike all recent mobile-efficient NAS methods which only 
release pretrained models, we open-source our entire codebase at:
\url{https://github.com/dstamoulis/single-path-nas}.

\end{abstract}

\section{Introduction}

``\textit{Is it possible to reduce the considerable search cost of Neural 
Architecture Search (NAS) down to only \textit{few hours}?}''
NAS has revolutionized the design of Convolutional Networks
(ConvNets)~\cite{zoph2017learning},
yielding state-of-the-art results in several deep learning applications~\cite{real2018regularized}.
NAS methods already have a profound impact on the design of hardware-efficient 
ConvNets for computer vision tasks under the constraints (\textit{e.g.}, 
inference latency) imposed by mobile devices~\cite{tan2018mnasnet}.

Despite the recent breakthroughs, NAS remains an intrinsically costly 
optimization problem. Searching for 
which convolution operation to use per ConvNet layer, gives rise to a 
combinatorially large search space: \textit{e.g.}, for a mobile-efficient 
ConvNet with 22 layers, choosing among five candidate operations
yields $5^{22} \approx 10^{15}$ possible ConvNet architectures. 
To traverse this design space, earlier NAS methods guide the exploration
via reinforcement learning (RL)~\cite{tan2018mnasnet}. Nonetheless, 
training the RL controller poses prohibitive computational challenges, 
and thousands of candidate ConvNets need to be trained~\cite{wu2018fbnet}.

\begin{figure}[ht!]
  \centering
  \includegraphics[width=1.0\columnwidth]{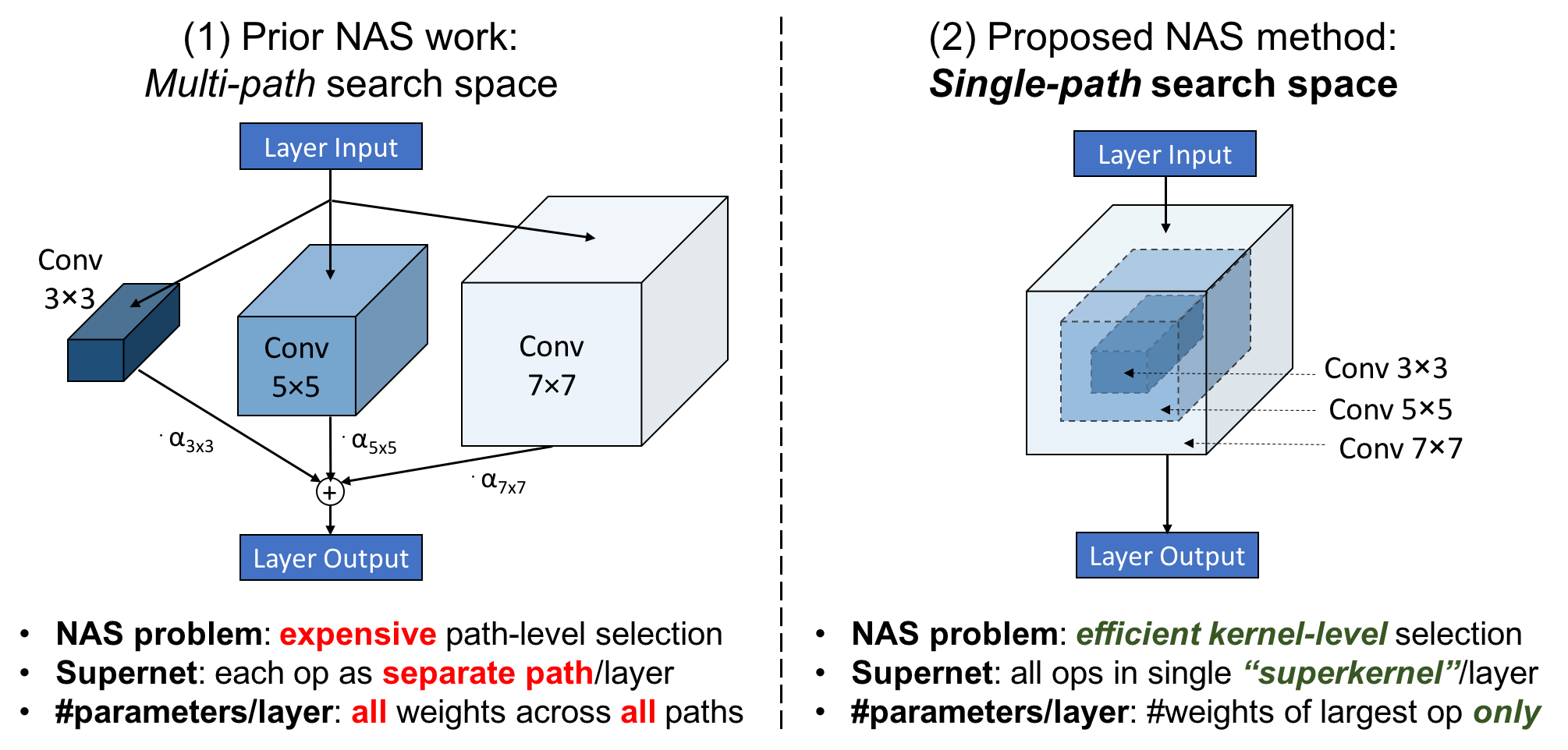}
  \caption{\textit{Single-Path NAS} directly 
  optimizes for the subset of convolution kernel weights and searches
  over an over-parameterized ``\textbf{superkernel}'' in each ConvNet layer
  (right). This \textbf{novel view} of the design space
  eliminates the need for maintaining separate paths for each candidate 
  operation, as in previous \textit{multi-path} approaches (left). 
  Our \textbf{key insight} drastically reduces the NAS search cost by up to 
  \textbf{5,000$\times$} with state-of-the-art accuracy 
  on ImageNet for the same mobile latency setting, compared to prior work.}
  \label{fig:key_idea}
\end{figure}

\textbf{Inefficiencies of \textit{multi-path} NAS}:
Recent NAS literature has seen a shift towards one-shot differentiable
formulations~\cite{liu2018darts,pham2018efficient,xie2018snas}
which search over a supernet that encompasses all candidate 
architectures. Specifically, current NAS methods relax 
the combinatorial optimization problem of finding the optimal ConvNet 
architecture to an operation/path selection problem: first, 
an over-parameterized, \textit{multi-path} supernet is constructed, 
where, for each layer, every candidate operation is added as a 
\textit{separate} trainable path, as illustrated in Figure~\ref{fig:key_idea} (left).
Next, NAS formulations solve for the (distributions of) paths of the 
\textit{multi-path} supernet that yield the optimal architecture.

As expected, naively branching out all paths is inefficient 
due to an intrinsic limitation: the number of trainable 
parameters that need to be maintained and updated 
during the search grows linearly with respect to the number 
of candidate operations per layer~\cite{bender2018understanding}.
To tame the memory explosion introduced by the \textit{multi-path} supernet,
current methods employ creative ``workaround'' solutions:
\textit{e.g.}, searching on a proxy dataset (subset of ImageNet~\cite{wu2018fbnet}), 
or employing a memory-wise scheme with only a subset of paths being updated during 
the search~\cite{cai2018proxylessnas}. Nevertheless, these techniques 
remain considerably costly, with an overall computational demand of
at least 200 GPU-hours.

In this paper, we propose \textit{Single-Path NAS}, a novel NAS method for designing
hardware-efficient ConvNets in \textbf{less than 4 hours}. Our \textbf{key insight} 
is illustrated in Figure~\ref{fig:key_idea} (right). We build upon the 
observation that different candidate convolutional operations in NAS 
can be viewed as subsets of a \textbf{single ``superkernel''}. Without having to 
choose among different paths/operations as in \textit{multi-path} methods, we instead 
solve the NAS problem as \textit{finding which subset of kernel weights to use 
in each ConvNet layer}. By sharing the convolutional kernel weights, 
we encode all candidate NAS operations into a single \textbf{``superkernel''}, 
\textit{i.e.}, with a single path, for each layer of the one-shot NAS supernet. 
This novel encoding of the design space yields a drastic reduction to 
the number of trainable parameters/gradients, allowing our NAS method to use 
batch sizes of $1024$, a four-fold increase compared to prior art's 
search efficiency.

Our contributions are as follows:
\begin{enumerate}

\item \textbf{Single-path NAS}: We propose a novel 
view of the one-shot, supernet-based design space, hence drastically 
decreasing the number of trainable parameters. To the best of our knowledge,
this is the \textit{first} work to formulate the NAS problem 
as finding the subset of kernel weights in each ConvNet layer.

\item  \textbf{State-of-the-art results}: \textit{Single-Path NAS} achieves 
$74.96\%$ top-1 accuracy on ImageNet with 79ms latency on a Pixel 1, 
\textit{i.e.}, a $+0.31\%$ improvement over the current best
hardware-aware NAS~\cite{tan2018mnasnet} under $80ms$.

\item  \textbf{NAS efficiency}: The overall search cost is only 
\textbf{8 epochs}, \textit{i.e.}, \textbf{3.75 hours} on TPUs  
(30 TPU-hours), up to \textbf{5,000$\times$ faster} compared to prior work. 

\item  \textbf{Reproducibility}: Unlike recent hardware-efficient 
NAS methods which release pretrained models only, we open-source and fully document
our method at: \url{https://github.com/dstamoulis/single-path-nas}.

\end{enumerate}

\section{Related Work}

\textbf{Hardware-efficient ConvNets}: While complex ConvNet designs
have unlocked unprecedented performance levels in computer vision tasks,
the accuracy improvement has come at the cost of higher computational 
complexity, making the deployment of state-of-the-art ConvNets to 
mobile devices challenging~\cite{stamoulis2018designing}. To this end, 
a significant body of prior work aims to co-optimize
for the inference latency of ConvNets. Earlier approaches focus on human expertise
to introduce hardware-efficient
operations~\cite{howard2017mobilenets,sandler2018mobilenetv2,zhang1707shufflenet}.
Pruning~\cite{chin2018layer} and quantization~\cite{ding2017lightnn} methods 
share the same goal to improve the efficiency of ConvNets.

\textbf{Neural Architecture Search (NAS)}: NAS aims at 
automating the process of designing ConvNets, giving rise to 
methods based on reinforcement learning (RL), evolutionary 
algorithms, or gradient-based 
methods~\cite{liu2018darts,pham2018efficient,real2018regularized,zoph2016neural,zoph2017learning}.
Earlier approaches train an agent (\textit{e.g.}, RNN controller)
by sampling candidate architectures over a cell-based 
design space, where the same cell is repeated in all layers
and the focus is on searching the cell architecture~\cite{zoph2017learning}. 
Nonetheless, training the controller
over different architectures makes the search costly. 

\textbf{Hardware-aware NAS}: Earlier NAS methods focused 
on maximizing accuracy under FLOPs constraints~\cite{xie2018snas,zhou2018resource}, 
but low FLOP count does not necessarily translate to hardware 
efficiency~\cite{dong2018dpp,stamoulis2018hyperpower}.
More recent methods incorporate hardware terms (\textit{e.g.}, runtime, power)
into cell-based NAS formulations~\cite{dong2018dpp,hsu2018monas}, but 
cell-based implementations are not hardware friendly~\cite{wu2018fbnet}.
Breaking away from cell-based assumptions in the search space encoding, recent work 
employs NAS over a generalized MobileNetV2-based design space 
introduced in~\cite{tan2018mnasnet}.

\textbf{Hardware-aware Differentiable NAS}: 
Recent NAS literature has seen a shift towards one-shot 
NAS formulations~\cite{liu2018darts,pham2018efficient,xie2018snas}. 
Gradient-based NAS in particular has gained increased popularity and 
has achieved state-of-the-art results~\cite{brock2017smash}. One-shot-based 
methods use an over-parameterized super-model network, where, for each layer, every 
candidate operation is added as a separate trainable path.
Nonetheless, \textit{multi-path} 
search spaces have an intrinsic limitation: the number of trainable 
parameters that need to be maintained and updated with gradients 
during the search grows linearly with respect to the number of different
convolutional operations per layer, resulting in memory 
explosion~\cite{bender2018understanding,cai2018proxylessnas}.

To this end, state-of-the-art approaches employ different novel ``workaround'' solutions.
FBNet~\cite{wu2018fbnet} searches on a ``proxy'' dataset (\textit{i.e.},
subset of the ImageNet dataset). Despite the decreased search cost thanks 
to the reduced number of training images, these approaches do not address the fact 
that the entire supermodel needs to be maintained in memory during search, 
hence the efficiency is limited due to inevitable use of smaller batch sizes. 
ProxylessNAS~\cite{cai2018proxylessnas} has employed a memory-wise one-shot model 
scheme, where only a set of paths is updated during the search. However, such implementation-wise 
improvements do not address a second key suboptimality of one-shot approaches, 
\textit{i.e.}, the fact that separate gradient steps are needed to update the 
weights and the architectural decisions interchangeably~\cite{liu2018darts}. Although the 
number of trainable parameters, with respect to the memory cost, is kept to the same 
level at any step, the way that \textit{multi-path}-based methods traverse the 
design space remains inefficient.

\section{Proposed Method: \textit{Single-Path} NAS}

In this Section, we present our proposed 
method. First, we discuss our novel \textit{single-path} view 
(Subsection~\ref{subsec:view}) of the search space. Next, 
we encode the NAS problem as finding the 
subset of convolution weights over the \textit{over-parameterized} 
\textbf{``superkernel''} (Subsection~\ref{subsec:single-path-kernel}), and we discuss 
how it compares to existing \textit{multi-path}-based NAS 
(Subsection~\ref{subsec:comparison-vs-multi}). Last, we formulate 
the hardware-aware NAS objective function, where we incorporate 
an accurate inference latency model of ConvNets executing on the
Pixel~1 smartphone (Subsection~\ref{subsec:hw-loss}).

\begin{figure}[ht!]
  \centering
  \includegraphics[width=1.0\columnwidth]{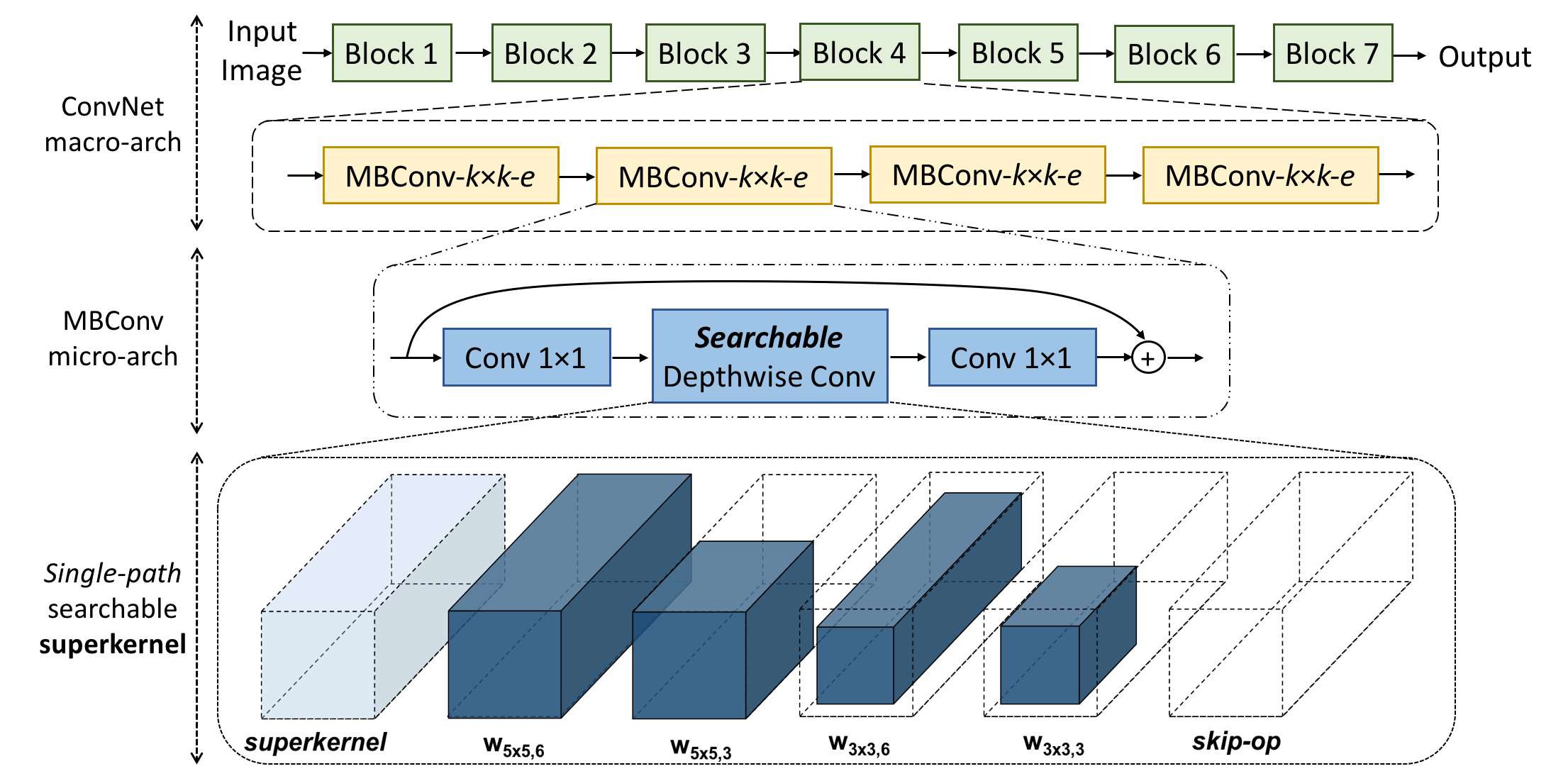}
  \caption{\textbf{\textit{Single-path} search space}: Our method builds upon
  \textit{hierarchical} MobileNetV2-like search 
  spaces~\cite{sandler2018mobilenetv2,tan2018mnasnet}, where the goal
  is to identify the type of mobile inverted bottleneck 
  convolution (MBConv)~\cite{sandler2018mobilenetv2} per layer.
  Our \textit{one-shot supernet} encapsulates all possible NAS architectures 
  in the search space, without the need for appending each candidate operation
  as a separate path. \textit{Single-Path} NAS directly searches over the weights of 
  a \textbf{searchable ``superkernel''} that encodes all MBConv types.}
  \label{fig:design_space}
\end{figure}

\subsection{Mobile ConvNets Search Space: A Novel View}
\label{subsec:view}

\textbf{Background - Mobile ConvNets}: State-of-the-art  
NAS builds upon a fixed ``backbone'' ConvNet~\cite{cai2018proxylessnas} inspired 
by the MobileNetV2 design~\cite{sandler2018mobilenetv2}, illustrated in 
Figure~\ref{fig:design_space} (top). Specifically, in this fixed macro-architecture, 
except for the head and stem layers, all ConvNet layers are grouped into 
blocks based on their filter sizes. The filter numbers per block 
follow the values in~\cite{wu2018fbnet}, \textit{i.e.}, we use 
seven blocks with up to four layers each. Each layer of these blocks 
follows a mobile inverted bottleneck convolution MBConv~\cite{sandler2018mobilenetv2}
micro-architecture, which consists of a point-wise ($1\times 1$) convolution, a $k\times k$ 
depthwise convolution, and a linear $1\times 1$ convolution (Figure~\ref{fig:design_space}, middle). 
Unless the layer has a stride 
value of two, a skip path is introduced to provide a residual 
connection from input to output. 

Each MBConv layer is parameterized by $k$, \textit{i.e.}, the kernel size
of the depthwise convolution, and by expansion ratio $e$, \textit{i.e.}, 
the ratio between the output and input of the first 
$1\times 1$ convolution. Based on this parameterization, we denote 
each MBConv as MBConv-$k\times k$-$e$.
Mobile-efficient NAS aims to choose each MBConv-$k\times k$-$e$ layer, 
by selecting among different $k$ and $e$ values~\cite{cai2018proxylessnas,wu2018fbnet}. 
In particular, we consider MBConv layers with kernel sizes $\{3,5\}$ and 
expansion ratios $\{3,6\}$. NAS also considers a special 
\textit{skip-op} ``layer'', which ``zeroes-out'' the kernel and feeds 
the input directly to the output, \textit{i.e.}, the entire layer is dropped.

\textbf{Novel view of design space}: 
Our \textit{key insight} is illustrated in Figure~\ref{fig:design_space}. 
We build upon the observation that different candidate convolutional operations in NAS 
can be viewed as subsets of the weights of an over-parameterized 
\textbf{single ``superkernel''} (Figure~\ref{fig:design_space}, bottom). 
This observation allows us to view the NAS combinatorial problem as 
\textit{finding which subset of kernel weights to use in each MBConv layer}. 
This observation is important since it allows
sharing the kernel parameters across different MBConv
architectural options. As shown in Figure~\ref{fig:design_space},
we encode all candidate NAS operations to this single 
\textbf{``superkernel''}, \textit{i.e.}, with a \textbf{single path}, for each
layer of the one-shot NAS supernet.

\subsection{Proposed Methodology: Single-Path NAS formulation}
\label{subsec:single-path-kernel}

\begin{figure}[h!]
  \centering
  \includegraphics[width=1.0\columnwidth]{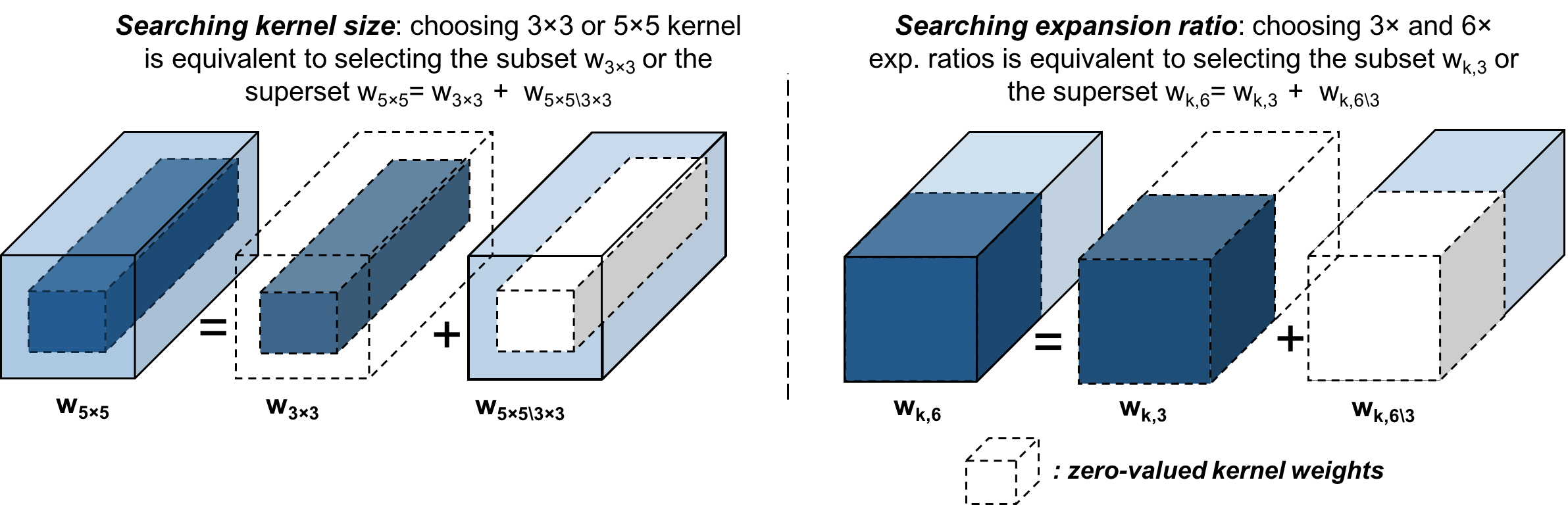}
  \caption{Encoding NAS decisions into the \textbf{superkernel}: 
  We formulate all candidate convolution operations (\textit{i.e.},
  different kernel size (left) and expansion ratio (right) values) directly 
  into the \textbf{searchable superkernel}.}
  \label{fig:key_idea_2}
\end{figure}

\textbf{Key idea - Relaxing NAS decisions over an over-parameterized kernel}: 
To simplify notation and to illustrate the key idea, 
without loss of generality, we show the case of choosing between a 
$3 \times 3$ or a $5 \times 5$ kernel for an MBConv layer.
Let us denote the weights of the two candidate kernels as 
$\textbf{w}_{3 \times 3}$ and $\textbf{w}_{5 \times 5}$, respectively. 
As shown in Figure~\ref{fig:key_idea_2} (left), we observe that 
the weights of the $3 \times 3$ kernel can be viewed as 
the \textit{inner} core of the weights of the $5 \times 5$ kernel, 
while ``zeroing'' out the weights of the ``\textit{outer}'' shell.
We denote this (\textit{outer}) subset of weights (that does not contribute 
to output of the $3 \times 3$ kernel but only to the $5 \times 5$ kernel), 
as $ \textbf{w}_{5 \times 5 \setminus 3 \times 3}$.
Hence, the NAS architectural choice of using 
the $5 \times 5$ convolution corresponds to using both 
the \textit{inner} $\textbf{w}_{3 \times 3}$ weights and the \textit{outer} shell, 
\textit{i.e.}, $\textbf{w}_{5 \times 5} = \textbf{w}_{3 \times 3} + 
\textbf{w}_{5 \times 5 \setminus 3 \times 3}$ (Figure~\ref{fig:key_idea_2}, left). 

We can therefore encode the NAS decision directly into the 
\textbf{superkernel} of an MBConv layer as a function of kernel weights
as follows:
\begin{equation}
    \label{eq:idea}
    \textbf{w}_{k} =  \textbf{w}_{3 \times 3} + \mathbbm{1}(\text{use~} 5 \times 5)  
    \cdot \textbf{w}_{5 \times 5 \setminus 3 \times 3}
\end{equation}
where $\mathbbm{1}(\cdot)$ is the indicator function that encodes the 
architectural NAS choice, \textit{i.e.}, if $\mathbbm{1}(\cdot) = 1$ then 
$\textbf{w}_{k} = \textbf{w}_{3 \times 3} + \textbf{w}_{5 \times 5 \setminus 3 \times 3} = 
\textbf{w}_{5 \times 5} $, else $\mathbbm{1}(\cdot) = 0$ then  
$\textbf{w}_{k} = \textbf{w}_{3 \times 3} $.

\textbf{Trainable indicator/condition function}: While the indicator function
encodes the NAS decision, a critical choice is how to formulate the condition over 
which the $\mathbbm{1}(\cdot)$ is evaluated. Our intuition is that, for an 
indicator function that represents whether to use the subset of weights,
its condition should be \textit{directly a function of the subset's weights}. 
Thus, our goal is to define an ``importance'' signal of the subset
weights that intrinsically captures their contribution to the overall ConvNet
loss. We draw inspiration from weight-based conditions that have been 
successfully used for quantization-related 
decisions~\cite{ding2019flightnns} and we use 
the \textit{group Lasso term}. Specifically, for the indicator related to
the $\textbf{w}_{5 \times 5 \setminus 3 \times 3}$ ``outer shell'' decision, 
we write the following condition:
\begin{equation}
    \label{eq:proposed-form-2}
    \textbf{w}_{k} = \textbf{w}_{3 \times 3} + \mathbbm{1}(\left\Vert 
    \textbf{w}_{5 \times 5 \setminus 3 \times 3} \right\Vert^2 > t_{k=5}) 
    \cdot \textbf{w}_{5 \times 5 \setminus 3 \times 3}
\end{equation}
where $t_{k=5}$ is a latent variable that controls the decision (\textit{e.g.},
a threshold value) of selecting kernel $5 \times 5$. The threshold will be compared to 
the Lasso term to determine if the \textit{outer}
$\textbf{w}_{5 \times 5 \setminus 3 \times 3 }$ weights are used to the overall convolution.
It is important to notice that, instead of picking the thresholds (\textit{e.g.}, $t_{k=5}$) 
by hand, we seamlessly 
treat them as trainable parameters to learn via gradient descent. 
To compute the gradients for thresholds, we relax the indicator 
function $g(x,t) = \mathbbm{1}(x>t)$ to a 
sigmoid function, $\sigma(\cdot)$, when computing gradients, \textit{i.e.}, 
$\hat{g}(x,t) = \sigma(x>t)$.

\textbf{Searching for expansion ratio and skip-op}: Since the result of the 
kernel-based NAS decision $\textbf{w}_{k}$ (Equation~\ref{eq:proposed-form-2}) is a 
convolution kernel itself, we can in turn apply our formulation to also encode 
NAS decisions for the expansion ratio of the $\textbf{w}_{k}$ kernel.
As illustrated in Figure~\ref{fig:key_idea_2} (right), the channels of the 
depthwise convolution in an MBConv-$k\times k$-$3$ layer with expansion ratio $e=3$ 
can be viewed as using one half of the channels of an 
MBConv-$k\times k$-$6$ layer with expansion ratio $e=6$, while ``zeroing'' 
out the second half of channels $\{\textbf{w}_{k, 6 \setminus 3}\}$.
Finally, by ``zeroing'' out the first half of the output filters as well, 
the entire \textbf{superkernel} contributes nothing if added to the 
residual connection of the MBConv layer: \textit{i.e.}, by deciding if $e=3$,
we can encode the NAS decision of using, or not, only the ``skip-op'' path. 
For both decisions over $\textbf{w}_{k}$ kernel, we write:
\begin{equation}
    \label{eq:proposed-form-3}
    \textbf{w} = \mathbbm{1}(\left\Vert 
    \textbf{w}_{k, 3} \right\Vert^2 > t_{e=3}) \cdot 
    ( \textbf{w}_{k, 3} + \mathbbm{1}(\left\Vert 
    \textbf{w}_{k, 6 \setminus 3} \right\Vert^2 > t_{e=6}) 
    \cdot \textbf{w}_{k, 6 \setminus 3} )
\end{equation}
Hence, for  input $\textbf{x}$, the output of the $i$-th MBConv layer of the network is: 
\begin{equation}
    \label{eq:effective-kernel-output}
    o^i (\textbf{x}) = \text{conv}(\textbf{x}, \textbf{w}^{i} | t_{k=5}^{i}, t_{e=6}^{i}, t_{e=3}^{i})
\end{equation}

\textbf{Searchable MBConv kernels}: Each MBConv uses $1\times 1$ convolutions for 
the point-wise (first) and linear stages, while the kernel-size decisions 
affect only the (middle) $k\times k$ depthwise convolution (Figure~\ref{fig:design_space}). 
To this end, we use our \textbf{searchable} $k\times k$ depthwise 
kernel at this middle stage. In terms of number of channels, the depthwise 
kernel depends on the point-wise $1\times 1$ output, which allows us to 
directly encode the expansion ratio $e$ at the middle stage as well: 
by setting the point-wise $1\times 1$ output to 
the maximum candidate expansion ratio, we can instead solve for which 
of them not to ``zero'' out at the depthwise (middle) state.
In other words, we directly use our \textbf{searchable} depthwise convolution 
\textbf{superkernel} to effectively encode the NAS decision for the expansion ratio.
Hence, our \textit{single-path}, convolution-based formulation can sufficiently capture 
any MBConv type (\textit{e.g.}, MBConv-$3\times 3$-$6$, MBConv-$5\times 5$-$3$,
\textit{etc.}) in the MobileNetV2-based design space (Figure~\ref{fig:design_space}).

\subsection{Single-Path vs. Existing Multi-Path Assumptions}
\label{subsec:comparison-vs-multi}

\textbf{Comparison with multi-path over-parameterized network}:
We briefly illustrate how our \textit{single-path} formulation compares to multi-path 
NAS approaches. In existing methods~\cite{cai2018proxylessnas,liu2018darts,wu2018fbnet},
the output of each layer $i$ is a (weighted) sum defined over the output of $N$ 
different paths, where each path $j$ corresponds to a different candidate 
kernel $\textbf{w}^{i,j}_{k \times k, e}$. The weight of each path 
$\alpha^{i,j}$ corresponds to the probability that this path 
is selected over the parallel paths:
\begin{equation}
    \label{eq:comparison}
    o^{i}_{multi-path}(\textbf{x}) = \sum_{j=1}^N \alpha^{i,j} \cdot o^{i,j}(\textbf{x}) = 
    \alpha^{i,0} \cdot \text{conv}(\textbf{x}, \textbf{w}^{i,0}_{3 \times 3}) + \dots + \alpha^{i,N} 
    \cdot \text{conv}(\textbf{x}, \textbf{w}^{i,N}_{5 \times 5})
\end{equation}
It is easy to see how our novel \textit{single-path} view is advantageous, 
since the output of the convolution at layer $i$ of our search space 
is \textit{directly a function of the weights of our single over-parameterized kernel} 
(Equation~\ref{eq:effective-kernel-output}): 
\begin{equation}
    \label{eq:single-path-conv}
    o^{i}_{single-path}(\textbf{x}) = o^i (\textbf{x}) = \text{conv}(\textbf{x}, \textbf{w}^{i} | 
    t_{k=5}^{i}, t_{e=6}^{i}, t_{e=3}^{i})
\end{equation}

\textbf{Comparison with multi-path NAS optimization}: 
Multi-path NAS methods solve for the optimal architecture parameters 
$\alpha$ (path weights), such that the weights 
$w_{\alpha}$ of the corresponding $\alpha$-architecture have minimal 
loss $\mathcal{L} (\alpha, w_{\alpha})$:
\begin{equation}
    \label{eq:bilevel}
    \underset{\alpha}{\text{min }} \underset{w_{\alpha}}{\text{min }} \mathcal{L}(\alpha, w_{\alpha})
\end{equation}
However, solving Equation~\ref{eq:bilevel} gives rise to a challenging \textit{bi-level} 
optimization problem~\cite{liu2018darts}. Existing methods interchangeably 
update the $\alpha$'s while freezing the $w$'s and vice versa, leading to more gradient steps.

In contrast, with our \textit{single-path} formulation, the overall network loss
is directly a function of the \textbf{``superkernel''} weights, where the learnable
kernel- and expansion ratio-related threshold variables, $\textbf{t}_k$
and $\textbf{t}_e$, are directly derived as a function (norm) of
the kernel weights $\textbf{w}$. Consequently, \textit{Single-Path NAS} formulates 
the NAS problem as solving \textit{directly over the weight kernels $\textbf{w}$ 
of a single-path, compact neural network}. Formally, the NAS problem becomes:
\begin{equation}
    \label{eq:sp-nas}
    \underset{\textbf{w}}{\text{min }} \mathcal{L}(\textbf{w} | \textbf{t}_{k}, \textbf{t}_{e})
\end{equation}

\textbf{Efficiency of \textit{Single-Path NAS}}: 
Unlike the bi-level optimization problem in prior work, solving
our NAS formulation in Equation~\ref{eq:sp-nas} is as expensive as
training the weights of a single-path, \textbf{branchless}, compact neural network
with vanilla gradient descent. Therefore, our formulation eliminates the need 
for separate gradient steps between the ConvNet
weights and the NAS parameters. Moreover, the reduction of the trainable 
parameters $\textbf{w}$ per se, further leads to a drastic reduction of
the search cost down to \textbf{just a few epochs},
as our experimental results show later in Section~\ref{sec:results}.
Our NAS problem formulation allows us to efficiently 
solve Equation~\ref{eq:sp-nas} with batch sizes of 1024, 
a four-fold increase compared to prior art's search efficiency.

\subsection{Hardware-Aware NAS with Differentiable Runtime Loss}
\label{subsec:hw-loss}

To design hardware-efficient ConvNets, the differentiable objective in Equation~\ref{eq:sp-nas}
should reflect both the accuracy of the searched architecture and its inference latency
on the target hardware. Hence, we use a latency-aware 
formulation~\cite{cai2018proxylessnas,wu2018fbnet}:
\begin{equation}
    \label{eq:loss}
    \mathcal{L}(\textbf{w} | \textbf{t}_{k}, \textbf{t}_{e} ) = 
    CE(\textbf{w} | \textbf{t}_{k}, \textbf{t}_{e} ) + \lambda \cdot 
    \text{log}(R(\textbf{w} | \textbf{t}_{k}, \textbf{t}_{e}))
\end{equation}
The first term $CE$ corresponds to the cross-entropy loss of the
single-path model. The hardware-related term $R$ is the 
runtime in milliseconds ($ms$) of the searched NAS model on the target 
mobile platform. Finally, the coefficient $\lambda$ modulates
the trade-off between cross-entropy and runtime.

\textbf{Runtime model over the single-path design space}: To preserve the
differentiability of the objective, another critical choice is the formulation of
the latency term $R$. Prior art has showed that the total network latency of a 
mobile ConvNet can be modeled as the sum of each $i$-th layer's 
runtime $R^{i}$, since the runtime of each operator
is independent of other operators~\cite{cai2017neuralpower,cai2018proxylessnas,wu2018fbnet}:
\begin{equation}
    \label{eq:runtime-network}
    R(\textbf{w} | \textbf{t}_{k}, \textbf{t}_{e}) = 
    \sum_{i} R^{i}(\textbf{w}^{i} | \textbf{t}^{i}_{k}, \textbf{t}^{i}_{e})
\end{equation}

For our approach, we adapt the per-layer runtime model as a function of the 
NAS-related decisions $\textbf{t}$.
We profile the target mobile platform (Pixel 1)
and we record the runtime for each candidate kernel operation per layer $i$, 
\textit{i.e.}, $R^{i}_{3 \times 3,3}$, $R^{i}_{3 \times 3,6}$, $R^{i}_{5 \times 5,3}$,
and $R^{i}_{5 \times 5,6}$. We denote the runtime of layer $i$ by following the notation 
in Equation~\ref{eq:proposed-form-3}. Specifically, 
the runtime of layer $i$ is defined first as a
function of the expansion ratio decision:
\begin{equation}
    \label{eq:runtime-layer-e}
    R^{i}_{e} = \mathbbm{1}(\left\Vert 
    \textbf{w}_{k, 3} \right\Vert^2 > \textbf{t}_{e=3}) \cdot 
    ( R^{i}_{5 \times 5, 3} + \mathbbm{1}(\left\Vert 
    \textbf{w}_{k, 6 \setminus 3} \right\Vert^2 > \textbf{t}_{e=6}) 
    \cdot ( R^{i}_{5 \times 5, 6} - R^{i}_{5 \times 5, 3} ))
\end{equation}
Next, by incorporating the kernel size decision, the total runtime is:
\begin{equation}
    \label{eq:runtime-layer}
    R^{i} = \frac{R^{i}_{3 \times 3,6}}{R^{i}_{5 \times 5,6}} \cdot R^{i}_{e} + 
    R^{i}_{e} \cdot (1-\frac{R^{i}_{3 \times 3,6}}{R^{i}_{5 \times 5,6}}) \cdot 
    \mathbbm{1}(\left\Vert 
    \textbf{w}_{5 \times 5 \setminus 3 \times 3} \right\Vert^2 > \textbf{t}_{k=5})
\end{equation}
As in Equation~\ref{eq:proposed-form-2}, we relax the indicator 
function to a sigmoid function $\sigma(\cdot)$ when computing gradients.
By using this model, the runtime term in the loss function remains 
differentiable with respect to layer-wise NAS choices.
As we show in our results, the model is accurate, with an average 
prediction error of $1.76\%$.

\section{Experiments}
\label{sec:results}

\subsection{Experimental Setup}

\textbf{Dataset and target application}:
We use \textit{Single-Path NAS} to design ConvNets for image classification 
on ImageNet~\cite{deng2009imagenet}. We use Pixel 1 as the target 
mobile platform. The choice of this experimental setup is important,
since it allows for a representative comparison with prior 
hardware-efficient NAS methods that optimize for the same
Pixel 1 device around a target latency of 
$80ms$~\cite{cai2018proxylessnas,tan2018mnasnet}.

\textbf{Implementation and deployment}: We implement our NAS 
framework in TensorFlow (\texttt{TF} version 1.12).
During both search and training stages, we use TPUs 
(version 2)~\cite{jouppi2017datacenter}. To this end,  
we build on top of the \texttt{TPUEstimator} classes following the 
TPU-related documentation of the MnasNet 
repository\footnote{\url{https://github.com/tensorflow/tpu/tree/master/models/official/mnasnet}}. 
Last, all models (ours and prior work) are deployed with 
TensorFlow TFLite to the mobile device. On the device, we profile 
runtime using the Facebook AI Performance Evaluation 
Platform (\texttt{FAI-PEP})\footnote{\url{https://github.com/facebook/FAI-PEP}}
that supports profiling for \texttt{tflite} models with detailed 
per-layer runtime breakdown.

\textbf{Implementing the custom ``superkernels''}: We use \texttt{Keras} 
to implement our trainable ``superkernels.'' Specifically, we define a custom
\texttt{Keras}-based depthwise convolution kernel where the output is a function
of both the weights and the threshold-based decisions 
(Equations~\ref{eq:proposed-form-2}-\ref{eq:proposed-form-3}). Our custom 
layer also returns the effective runtime of the layer 
(Equations~\ref{eq:runtime-layer-e}-\ref{eq:runtime-layer}). 
We document our implementation in our project GitHub 
repository: \url{https://github.com/dstamoulis/single-path-nas},
with detailed steps on how to reproduce the results.

\subsection{State-of-the-art Runtime-Constrained ImageNet Classification}

We apply our method to design ConvNets for the Pixel 1 phone
with an overall target latency of $80ms$. We train the derived \textit{Single-Path} 
NAS model for 350 epochs, following the MnasNet 
training schedule~\cite{tan2018mnasnet}. We compare our method with
mobile ConvNets designed by human experts and state-of-the-art 
NAS methods in Table~\ref{tab:imagenet-sota}, in terms of classification 
accuracy and search cost. In terms of hardware efficiency, prior work has 
shown that low FLOP count does 
not necessarily translate to high hardware efficiency~\cite{dong2018dpp}, 
we therefore evaluate the various NAS methods with respect to the inference
runtime on Pixel 1 ($\leq 80ms$).

\textbf{Enabling a representative comparison}:
While we provide the original values from the 
respective papers, our goal is to ensure a fair comparison. 
To this end, we retrain the baseline models following the same 
schedule (in fact, we find that the MnasNet-based training schedule 
improves the top1 accuracy compared to what is reported in several previous 
methods). Similarly, we profile the models on the same Pixel 1 device.
For prior work that does not optimize for Pixel 1, we retrain and profile their 
model closest to the MnasNet baseline (\textit{e.g.}, the FBNet-B and ChamNet-B 
networks~\cite{dai2018chamnet,wu2018fbnet}, since the authors use these
ConvNets to compare against the MnasNet model). Finally, to enable a representative 
comparison of the search cost per method, we directly report 
the number of epochs reported per method, hence canceling out the 
effect of different hardware systems (GPU vs TPU hours).

\textbf{ImageNet classification}:
Table~\ref{tab:imagenet-sota} shows that our \textit{Single-Path} NAS achieves 
top-1 accuracy of $\textbf{74.96\%}$, which is the new state-of-the-art ImageNet accuracy
among hardware-efficient NAS methods. More specifically, 
\textbf{our method achieves better top-1 accuracy than ProxylessNAS
by} $+\textbf{0.31}\%$, while maintaining on par target latency of $\leq 80ms$ on the 
same target mobile phone. \textit{Single-Path} NAS outperforms methods in 
this mobile latency range, \textit{i.e.}, better than MnasNet ($+0.35\%$), 
FBNet-B ($+0.86\%$), and MobileNetV2 ($+1.37\%$).

\begin{table}[t!]
\caption{\textit{Single-Path} NAS achieves state-of-the-art accuracy (\%) on ImageNet 
for similar mobile latency setting compared to previous 
NAS methods ($\leq 80 ms$ on Pixel 1), with up to 
$5,000 \times$ reduced search cost in terms of number of epochs. *The search cost
in epochs is estimated based on the claim~\cite{cai2018proxylessnas}
that ProxylessNAS is $200 \times$ faster than MnasNet. $\ddag$ChamNet does not detail
the model derived under runtime constraints~\cite{dai2018chamnet} 
so we cannot retrain or measure the latency.}
\centering
\scalebox{0.955}{
\begin{tabular}{l|cccc}
\hline
\multirow{2}{*}{Method} & Top-1 & Top-5 & Mobile & Search \\ 
  & Acc (\%) & Acc (\%) & Runtime (ms) & Cost (epochs)  \\ \hline \hline
  
MobileNetV1~\cite{howard2017mobilenets} & 70.60 & 89.50 & 113 & \multirow{3}{*}{-} \\
MobileNetV2 1.0x~\cite{sandler2018mobilenetv2} & 72.00 & 91.00 & 75.00 &  \\
MobileNetV2 1.0x (our impl.) & 73.59 & 91.41 & 73.57 &  \\\hline

Random search     & 73.78 $\pm$ 0.85 & 91.42 $\pm$ 0.56 & 77.31 $\pm$ 0.9 ms & - \\\hline


MnasNet 1.0x~\cite{tan2018mnasnet} & 74.00 & 91.80 & 76.00 & \multirow{2}{*}{40,000}  \\
MnasNet 1.0x (our impl.)  & 74.61 & 91.95 & 74.65 &   \\\hline
ChamNet-B~\cite{dai2018chamnet}   & 73.80 & -- & -- & 240$\ddag$  \\\hline
ProxylessNAS-R~\cite{cai2018proxylessnas} & 74.60 & 92.20 & 78.00 & \multirow{2}{*}{200*}  \\
ProxylessNAS-R (our impl.)  & 74.65 & 92.18 & 77.48 &   \\\hline
FBNet-B~\cite{wu2018fbnet} & 74.1 & - & - & \multirow{2}{*}{90}  \\
FBNet-B (our impl.)  & 73.70 & 91.51 & 78.33 &   \\\hline
\hline
\textit{Single-Path} NAS (\textbf{proposed}) & \textbf{74.96} & \textbf{92.21} & 79.48 & \textbf{8} (\textbf{3.75 hours})  \\\hline
\end{tabular}
}
\label{tab:imagenet-sota}
\end{table}

\textbf{NAS search cost}: \textit{Single-Path} NAS has \textbf{orders of magnitude 
reduced search cost} compared to all previous hardware-efficient NAS methods.
Specifically, MnasNet reports that the controller uses 8k sampled models, each 
trained for 5 epochs, for a total of 40k train epochs. In turn, ChamNet
trains an accuracy predictor on 240 samples, which assuming an aggressively 
fast training schedule of five epochs per sample (same as in MnasNet),
corresponds to a total search cost of 1.2k epochs.
ProxylessNAS reports $200 \times$ search cost improvement over MnasNet,
hence the overall cost is the TPU-equivalent of 200 epochs. 
Finally, FBNet reports 90 epochs of training on a proxy dataset (10\%
of ImageNet). While the number of images per epoch is reduced, we found
that a TPU can accommodate a FBNet-like supermodel with maximum batch size
of 128, hence the number of steps per FBNet epoch are still $8 \times$
more compared to the steps per epoch in our method.

\begin{figure}[h!]
  \centering
  \includegraphics[width=.5\columnwidth]{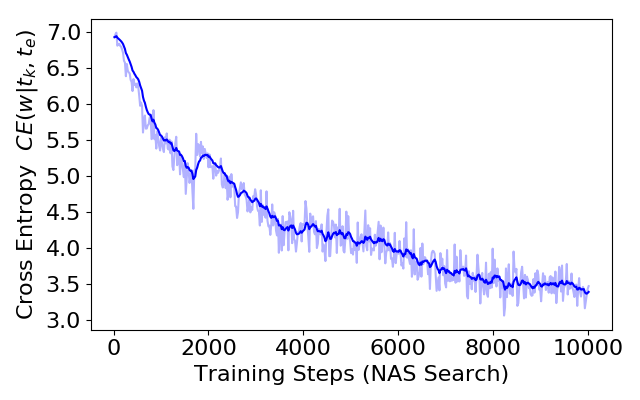}~
  \includegraphics[width=.5\columnwidth]{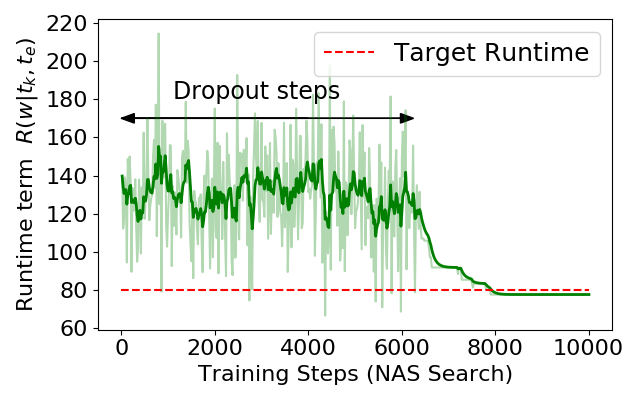}
  \caption{\textit{Single-Path NAS} search progress: Progress of both objective terms, 
  \textit{i.e.}, cross entropy $CE$ (left) and runtime $R$ (right) during NAS search.}
  \label{fig:progress}
\end{figure}

\begin{figure}[t]
  \centering
  \includegraphics[width=1.0\columnwidth]{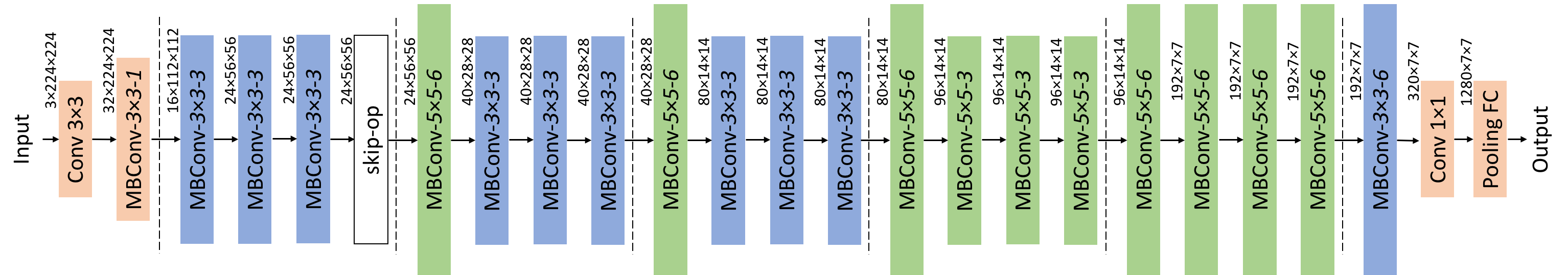}
  \caption{Hardware-efficient ConvNet found by \textit{Single-Path} NAS, with 
  top-1 accuracy of $\textbf{74.96\%}$ on ImageNet and inference time of $79.48 ms$ 
  on Pixel 1 phone.}
  \label{fig:spnet}
\end{figure}

In comparison, \textit{Single-Path NAS} has a total cost of eight epochs, which 
is $\textbf{5,000} \times$ faster than MnasNet, $\textbf{25} \times$ faster than ProxylessNAS, 
and $\textbf{11} \times$ faster than FBNet. In particular, we use an aggressive 
training schedule similar to the few-epochs schedule used in MnasNet to 
train the individual ConvNet samples~\cite{tan2018mnasnet}. Due to space 
limitations, we provide implementation details (\textit{e.g.}, label smoothing, learning rates,
$\lambda$ value, \textit{etc.}) in our project repository. Overall, we visualize the search 
efficiency of 
our method in Figure~\ref{fig:progress}, where we show the progress of both 
$CE$ and $R$ terms of Equation~\ref{eq:sp-nas}. Earlier during our search (first six epochs), 
we employ \textit{dropout} across the different subsets of the kernel weights
(Figure~\ref{fig:progress}, right).
Dropout is a common technique in NAS methods to prevent the supernet from 
learning as an ensemble. Unlike prior art that employs this technique over 
the separate paths of the \textit{multi-path} supernet, we directly 
drop randomly the subsets of the superkernel in our
\textit{single-path} search space. 
We search for \textbf{$\sim 10k$ steps} (8 epochs with a batch size of $1024$),
which corresponds to total wall-clock time of \textbf{3.75 hours} on a TPUv2.
In particular, given than a TPUv2 has 2 chips with 4 cores each, this 
corresponds to a total of 30 TPU-hours.

\begin{figure}[t]
  \centering
  \includegraphics[width=1.0\columnwidth]{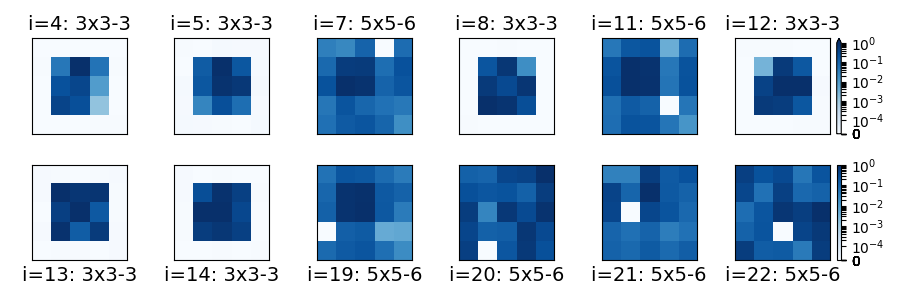}
  \caption{Visualization of kernel-based architectural contributions. The 
  \textit{standard deviation} of \textbf{superkernel} values across the kernel
  channels is shown in log-scale, with lighter colors indicating smaller values.}
  \label{fig:kernels}
\end{figure}

\textbf{Visualization of \textit{Single-Path NAS} ConvNet}: Our derived 
ConvNet architecture is shown in Figure~\ref{fig:spnet}. Moreover,
to illustrate how the \textbf{searchable superkernels} effectively 
capture NAS decisions across subsets of kernel weights,
we plot the standard deviation of weight values in 
Figure~\ref{fig:kernels} (shown in log-scale, with lighter colors 
indicating smaller values). Specifically, we compute the standard deviation 
of weights across the channel-dimension for all \textbf{superkernels}. 
For various layers shown in Figure~\ref{fig:kernels} (per $i$-th ConvNet's 
layer from Figure~\ref{fig:spnet}), we observe
that the \textit{outer} $\textbf{w}_{5 \times 5 \setminus 3 \times 3}$
``shells'' reflect the NAS architectural choices: for
layers where the entire $\textbf{w}_{5 \times 5}$ is selected, 
the $\textbf{w}_{5 \times 5 \setminus 3 \times 3}$ 
values drastically vary across the channels. On the contrary, 
for all layers where $3 \times 3$ convolution is selected, 
the \textit{outer} shell values do not vary significantly.

\textbf{Comparison with random search}: We find surprising that 
mobile-efficient NAS methods lack a comparison against random search. 
To this end, we randomly sample ten ConvNets based on our design space; we employ 
sampling by rejection, where we keep samples with predicted runtime
from $75ms$ to $80ms$. The average accuracy and runtime of the random samples
are reported in Table~\ref{tab:imagenet-sota}. We observe that,
while random search does not outperform NAS methods, the overall accuracy
is comparable to MobileNetV2. This highlights that the effectiveness of
NAS methods heavily relies upon the properties of the MobileNetV2-based 
design space. Nonetheless, the search cost of random search is not 
representative: to avoid training all ten samples,
we would follow a selection process similar to MnasNet, by training each 
sample for few epochs and picking the one with highest accuracy. 
Hence, the actual search cost for random search 
is not negligible, and for $\geq 10$ samples it is in fact comparable to 
automated NAS methods.

\begin{figure}[t]
    \centering
    \begin{tabular}{l  r}
    \begin{minipage}{0.46\linewidth}
        \includegraphics[width=\linewidth]{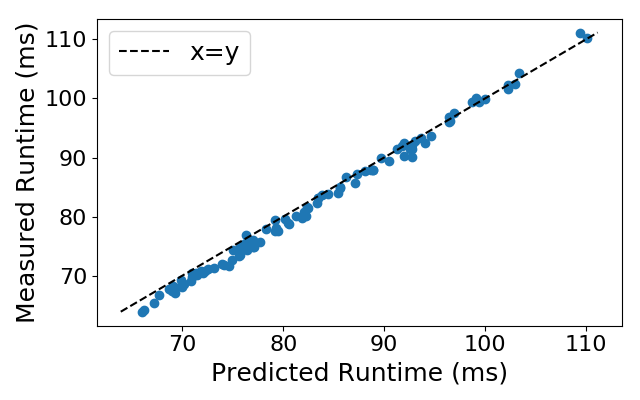}
        \caption{The runtime model (Equation~\ref{eq:runtime-network}) is accurate,
        with an average prediction error of $1.76\%$.}
        \label{fig:runtime_lut}
    \end{minipage}
    \qquad
    \begin{minipage}{0.46\linewidth}
        \includegraphics[width=\linewidth]{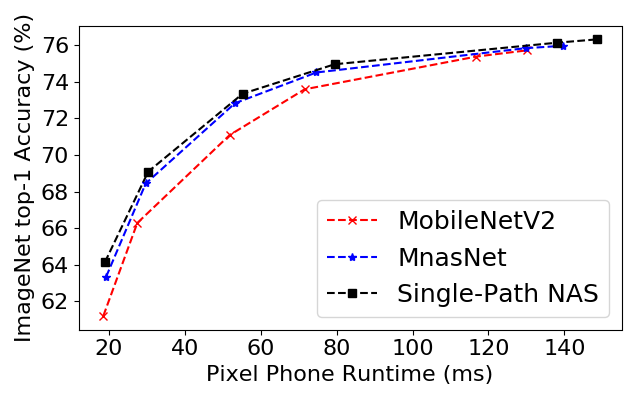}
        \caption{\textit{Single-Path} NAS outperforms MobileNetV2 and 
        MnasNet across various channel size scales.}
        \label{fig:depth_mult_figure}
    \end{minipage}
    \end{tabular}
\end{figure}

\textbf{Different channel size scaling}:
Next, we follow a typical analysis~\cite{cai2018proxylessnas,wu2018fbnet},
by rescaling the networks using a width multiplier~\cite{sandler2018mobilenetv2}.
As shown in Figure~\ref{fig:depth_mult_figure}, we observe 
that our model consistently outperforms prior methods under varying runtime 
settings. For instance, Single-Path NAS with $79.48ms$ is 1.56$\times$ faster
than the MobileNetV2 scaled model of similar accuracy.

\textbf{Runtime model}: To train the runtime model, we record the  
runtime per layer (MBConv operations breakdown) by profiling ConvNets with 
different MBConv types, \textit{i.e.}, we obtain the 
$R^{i}_{3 \times 3,3}$, $R^{i}_{3 \times 3,6}$, $R^{i}_{5 \times 5,3}$,
and $R^{i}_{5 \times 5,6}$ runtime values per MBConv layer $i$  
(Equations~\ref{eq:runtime-layer-e}-\ref{eq:runtime-layer}).
To evaluate the runtime-prediction accuracy of the model, 
we generate 100 randomly designed ConvNets and we measure their runtime
on the device. As illustrated in Figure~\ref{fig:runtime_lut},
our model can accurately predict the actual runtimes: 
the Root Mean Squared Error (RMSE) is $1.32ms$, which corresponds to an 
average $1.76\%$ prediction error. 

\begin{table}[t!]
\caption{Searching across subsets of kernel weights: 
ConvNets with weight values trained over subsets 
of the kernels ($3\times3$ as subset of $5 \times 5$)
achieve performance (top-1 accuracy) similar to ConvNets 
with individually trained kernels.}
\centering
\label{my-label}
\scalebox{0.9}{
\begin{tabular}{l|cc}
\hline
Method & Top-1  Acc (\%)  & Top-5  Acc (\%)   \\\hline\hline
  
Baseline ConvNet - $\textbf{w}_{3 \times 3} $ kernels & 73.59 & 91.41   \\
Baseline ConvNet - $\textbf{w}_{5 \times 5} $ kernels & 74.10 & 91.67   \\ \hline
\textit{Single-Path ConvNet} - inference w/ $\textbf{w}_{3 \times 3} $ kernels  & 73.43 & 91.42  \\
\textit{Single-Path ConvNet} - inference w/ $\textbf{w}_{3 \times 3}  + \textbf{w}_{5 \times 5 \setminus 3 \times 3} $ kernels & 73.86 & 91.72 \\
\hline
\end{tabular}
}
\label{tab:levels}
\end{table}

\subsection{Ablation Study: Kernel-based Accuracy-Efficiency Trade-off}

\textit{Single-Path NAS} searches over subsets of
the convolutional kernel weights. Hence, we conduct experiments to 
highlight how kernel-weight subsets can capture 
accuracy-efficiency trade-off effectively. To this end,
we use the MobileNetV2 macro-architecture as a backbone (we maintain the 
location of stride-2 layers as default). As two baseline networks, 
we consider the default MobileNetV2 with MBConv-$3\times 3$-$6$ blocks 
(\textit{i.e.}, $ \textbf{w}_{3 \times 3} $ kernels for all depthwise convolutions), 
and a network with MBConv-$5\times 5$-$6$ blocks (\textit{i.e.}, $ \textbf{w}_{5 \times 5} $ 
kernels).

Next, to capture the subset-based training of weights during a \textit{Single-Path} 
NAS search, we consider a \textit{ConvNet} with MBConv-$5\times 5$-$6$ blocks, where we 
compute the loss of the model over two subsets, 
(i) the inner $ \textbf{w}_{3 \times 3} $ weights, and (ii) by also using the 
remaining $\textbf{w}_{5 \times 5 \setminus 3 \times 3} $ 
weights. For each loss computed over these subsets, we accumulate back-propagated 
gradients and update the respective weights, \textit{i.e.}, gradients are being 
applied separately to the inner and to the entire kernel per layer. We
follow training steps similar to the ``switchable'' 
training across channels as in~\cite{yu2018slimmable} (for the remaining 
training hyper-parameters we use the same setup as the default MnasNet).
As shown in Table~\ref{tab:levels}, we observe the final accuracy
across the kernel granularity, \textit{i.e.}, with the 
inner $ \textbf{w}_{3 \times 3} $ and the 
entire $ \textbf{w}_{5 \times 5} =  \textbf{w}_{3 \times 3}  + 
\textbf{w}_{5 \times 5 \setminus 3 \times 3} $ kernels, 
follows an accuracy change relative to ConvNets with
individually trained kernels. 

Such finding is significant in the context of NAS, since choosing over 
subsets of kernels can effectively capture the accuracy-runtime trade-offs 
similar to their individually trained counterparts. We therefore conjecture that
our efficient \textbf{superkernel}-based design search can be flexibly adapted 
and benefit the guided search space exploration in other RL-based
NAS methods. Beyond the NAS literature, our finding is closely 
related to Slimmable networks~\cite{yu2018slimmable}. SlimmableNets
limit however their analysis across the channel dimension, and 
our work is the first to study trade-offs across the NAS kernel dimension.

\section{Conclusion}

In this paper, we proposed \textit{Single-Path NAS}, a NAS method
that reduces the search cost for designing hardware-efficient ConvNets 
to \textbf{less than 4 hours}. The key idea is to revisit the 
one-shot \textbf{supernet} design space with a novel
\textit{single-path} view, by formulating the NAS problem as 
\textit{finding which subset of kernel weights
to use} in each ConvNet layer. \textit{Single-Path NAS} achieved 
$74.96\%$ top-1 accuracy on ImageNet with 79ms latency on a Pixel 1 
phone, which is state-of-the-art accuracy with latency on-par 
with previous NAS methods ($\leq 80ms$). More importantly, 
we reduced the search cost of hardware-efficient NAS down 
to only \textbf{8 epochs} (30 TPU-hours), which is up to 
\textbf{5,000$\times$ faster} compared to prior work. 
\textbf{Impact beyond differentiable NAS}: 
While we used a differentiable NAS formulation, our novel design 
space encoding can be flexibly incorporated into other NAS methodologies. 
Hence, \textit{Single-Path NAS} could enable future work that 
builds upon the efficiency of our \textit{single-path}, one-shot 
design space for RL- or evolutionary-based NAS methods.

\section*{Acknowledgements}
This research was supported in part by National Science Foundation CSR 
Grant No. 1815780 and National Science Foundation CCF Grant No. 1815899. 
Dimitrios Stamoulis also acknowledges support from the Qualcomm Innovation 
Fellowship (QIF) 2018 and the TensorFlow Research Cloud programs.

%
%
\bibliographystyle{splncs04}
\bibliography{singlepathnas}

\begin{thebibliography}{10}
\providecommand{\url}[1]{\texttt{#1}}
\providecommand{\urlprefix}{URL }
\providecommand{\doi}[1]{https://doi.org/#1}

\bibitem{bender2018understanding}
Bender, G., Kindermans, P.J., Zoph, B., Vasudevan, V., Le, Q.: Understanding
  and simplifying one-shot architecture search. In: International Conference on
  Machine Learning. pp. 549--558 (2018)

\bibitem{brock2017smash}
Brock, A., Lim, T., Ritchie, J.M., Weston, N.: Smash: one-shot model
  architecture search through hypernetworks. arXiv preprint arXiv:1708.05344
  (2017)

\bibitem{cai2017neuralpower}
Cai, E., Juan, D.C., Stamoulis, D., Marculescu, D.: Neuralpower: Predict and
  deploy energy-efficient convolutional neural networks. In: Asian Conference
  on Machine Learning. pp. 622--637 (2017)

\bibitem{cai2018proxylessnas}
Cai, H., Zhu, L., Han, S.: Proxyless{NAS}: Direct neural architecture search on
  target task and hardware. In: International Conference on Learning
  Representations (2019)

\bibitem{chin2018layer}
Chin, T.W., Zhang, C., Marculescu, D.: Layer-compensated pruning for
  resource-constrained convolutional neural networks. arXiv preprint
  arXiv:1810.00518  (2018)

\bibitem{dai2018chamnet}
Dai, X., Zhang, P., Wu, B., Yin, H., Sun, F., Wang, Y., Dukhan, M., Hu, Y., Wu,
  Y., Jia, Y., et~al.: Chamnet: Towards efficient network design through
  platform-aware model adaptation. arXiv preprint arXiv:1812.08934  (2018)

\bibitem{deng2009imagenet}
Deng, J., Dong, W., Socher, R., Li, L.J., Li, K., Fei-Fei, L.: Imagenet: A
  large-scale hierarchical image database. In: 2009 IEEE conference on computer
  vision and pattern recognition. pp. 248--255. Ieee (2009)

\bibitem{ding2019flightnns}
Ding, R., Liu, Z., Chin, T.W., Marculescu, D., Blanton, R.: Flightnns:
  Lightweight quantized deep neural networks for fast and accurate inference.
  In: 2019 Design Automation Conference (DAC) (2019)

\bibitem{ding2017lightnn}
Ding, R., Liu, Z., Shi, R., Marculescu, D., Blanton, R.: Lightnn: Filling the
  gap between conventional deep neural networks and binarized networks. In:
  Proceedings of the on Great Lakes Symposium on VLSI 2017. pp. 35--40. ACM
  (2017)

\bibitem{dong2018dpp}
Dong, J.D., Cheng, A.C., Juan, D.C., Wei, W., Sun, M.: Dpp-net: Device-aware
  progressive search for pareto-optimal neural architectures. arXiv preprint
  arXiv:1806.08198  (2018)

\bibitem{howard2017mobilenets}
Howard, A.G., Zhu, M., Chen, B., Kalenichenko, D., Wang, W., Weyand, T.,
  Andreetto, M., Adam, H.: Mobilenets: Efficient convolutional neural networks
  for mobile vision applications. arXiv preprint arXiv:1704.04861  (2017)

\bibitem{hsu2018monas}
Hsu, C.H., Chang, S.H., Juan, D.C., Pan, J.Y., Chen, Y.T., Wei, W., Chang,
  S.C.: Monas: Multi-objective neural architecture search using reinforcement
  learning. arXiv preprint arXiv:1806.10332  (2018)

\bibitem{jouppi2017datacenter}
Jouppi, N.P., Young, C., Patil, N., Patterson, D., Agrawal, G., Bajwa, R.,
  Bates, S., Bhatia, S., Boden, N., Borchers, A., et~al.: In-datacenter
  performance analysis of a tensor processing unit. In: 2017 ACM/IEEE 44th
  Annual International Symposium on Computer Architecture (ISCA). pp. 1--12.
  IEEE (2017)

\bibitem{liu2018darts}
Liu, H., Simonyan, K., Yang, Y.: Darts: Differentiable architecture search.
  arXiv preprint arXiv:1806.09055  (2018)

\bibitem{pham2018efficient}
Pham, H., Guan, M.Y., Zoph, B., Le, Q.V., Dean, J.: Efficient neural
  architecture search via parameter sharing. arXiv preprint arXiv:1802.03268
  (2018)

\bibitem{real2018regularized}
Real, E., Aggarwal, A., Huang, Y., Le, Q.V.: Regularized evolution for image
  classifier architecture search. arXiv preprint arXiv:1802.01548  (2018)

\bibitem{sandler2018mobilenetv2}
Sandler, M., Howard, A., Zhu, M., Zhmoginov, A., Chen, L.C.: Mobilenetv2:
  Inverted residuals and linear bottlenecks. In: Proceedings of the IEEE
  Conference on Computer Vision and Pattern Recognition. pp. 4510--4520 (2018)

\bibitem{stamoulis2018hyperpower}
Stamoulis, D., Cai, E., Juan, D.C., Marculescu, D.: Hyperpower: Power-and
  memory-constrained hyper-parameter optimization for neural networks. In: 2018
  Design, Automation \& Test in Europe Conference \& Exhibition (DATE). IEEE
  (2018)

\bibitem{stamoulis2018designing}
Stamoulis, D., Chin, T.W.R., Prakash, A.K., Fang, H., Sajja, S., Bognar, M.,
  Marculescu, D.: Designing adaptive neural networks for energy-constrained
  image classification. In: Proceedings of the International Conference on
  Computer-Aided Design. ACM (2018)

\bibitem{tan2018mnasnet}
Tan, M., Chen, B., Pang, R., Vasudevan, V., Le, Q.V.: Mnasnet: Platform-aware
  neural architecture search for mobile. arXiv preprint arXiv:1807.11626
  (2018)

\bibitem{wu2018fbnet}
Wu, B., Dai, X., Zhang, P., Wang, Y., Sun, F., Wu, Y., Tian, Y., Vajda, P.,
  Jia, Y., Keutzer, K.: Fbnet: Hardware-aware efficient convnet design via
  differentiable neural architecture search. arXiv preprint arXiv:1812.03443
  (2018)

\bibitem{xie2018snas}
Xie, S., Zheng, H., Liu, C., Lin, L.: Snas: stochastic neural architecture
  search. arXiv preprint arXiv:1812.09926  (2018)

\bibitem{yu2018slimmable}
Yu, J., Yang, L., Xu, N., Yang, J., Huang, T.: Slimmable neural networks. arXiv
  preprint arXiv:1812.08928  (2018)

\bibitem{zhang1707shufflenet}
Zhang, X., Zhou, X., Lin, M., Sun, J.: Shufflenet: An extremely efficient
  convolutional neural network for mobile devices. arXiv preprint
  arXiv:1707.01083  (2017)

\bibitem{zhou2018resource}
Zhou, Y., Ebrahimi, S., Ar{\i}k, S.{\"O}., Yu, H., Liu, H., Diamos, G.:
  Resource-efficient neural architect. arXiv preprint arXiv:1806.07912  (2018)

\bibitem{zoph2016neural}
Zoph, B., Le, Q.V.: Neural architecture search with reinforcement learning.
  arXiv preprint arXiv:1611.01578  (2016)

\bibitem{zoph2017learning}
Zoph, B., Vasudevan, V., Shlens, J., Le, Q.V.: Learning transferable
  architectures for scalable image recognition. arXiv preprint arXiv:1707.07012
   \textbf{2}(6) (2017)

\end{thebibliography}

\end{document}